\documentclass{article}

\usepackage{microtype}
\usepackage{graphicx}
\usepackage{subfigure}
\usepackage{url}
\usepackage{amsmath}
\usepackage{amssymb}
\usepackage{multirow}
\usepackage{booktabs} 

\usepackage{hyperref}

\newcommand{\etal}{\textit{et al}.}

\newcommand{\eg}{\textit{e}.\textit{g}.}

\usepackage[accepted]{icml2021}

\begin{document}

\twocolumn[
\icmltitle{Performance Evaluation of Adversarial Attacks: Discrepancies and Solutions}



\icmlsetsymbol{equal}{*}

\begin{icmlauthorlist}
\icmlauthor{Jing Wu}{equal,UESTC}
\icmlauthor{Mingyi Zhou}{equal,UESTC}
\icmlauthor{Ce Zhu}{UESTC}
\icmlauthor{Yipeng Liu}{UESTC}
\icmlauthor{Mehrtash Harandi}{MN}
\icmlauthor{Li Li}{MN}
\end{icmlauthorlist}

\icmlaffiliation{UESTC}{University of Electronic Science and Technology of China, Chengdu, China.}
\icmlaffiliation{MN}{Monash University, Melbourne, Australia.}

\icmlcorrespondingauthor{Ce Zhu}{eczhu@uestc.edu.cn}

\icmlkeywords{Machine Learning, ICML}

\vskip 0.3in
]



\printAffiliationsAndNotice{\icmlEqualContribution} 

\begin{abstract}
Recently, adversarial attack methods have been developed to challenge the robustness of machine learning models. However, mainstream evaluation criteria experience limitations, even yielding discrepancies among results under different settings. By examining various attack algorithms, including gradient-based and query-based attacks, we notice the lack of a consensus on a uniform standard for unbiased performance evaluation. Accordingly, we propose a Piece-wise Sampling Curving (PSC) toolkit to effectively address the aforementioned discrepancy, by generating a comprehensive comparison among adversaries in a given range. In addition, the PSC toolkit offers options for balancing the computational cost and evaluation effectiveness. Experimental results demonstrate our PSC toolkit presents comprehensive comparisons of attack algorithms, significantly reducing discrepancies in practice. The codes are publicly available\footnote{{\color{blue} \url{https://github.com/AnonymousAuthor000/PSC}}}.
\end{abstract}

\section{Introduction}
Recent studies show deep neural networks are vulnerable to small perturbations on clean inputs, causing incorrect predictions \cite{szegedy2013intriguing}. According to this property of machine learning models, adversarial attacks have been proposed to evaluate the robustness of models. For evaluating gradient-based attacks, they usually compare attack success rate (ASR) under the same magnitude perturbation among different adversaries. For query-based attacks, they have three metrics to show, including the number of queries, the attack distance and ASR. In summary, current comparison methods for evaluating the performance of adversaries can be divided into two main categories, as shown in Table~\ref{current_standards}, includes point-wise comparison and curve-based comparison. 

\begin{figure*}[tbh]
\centering
\includegraphics[width=0.98\textwidth]{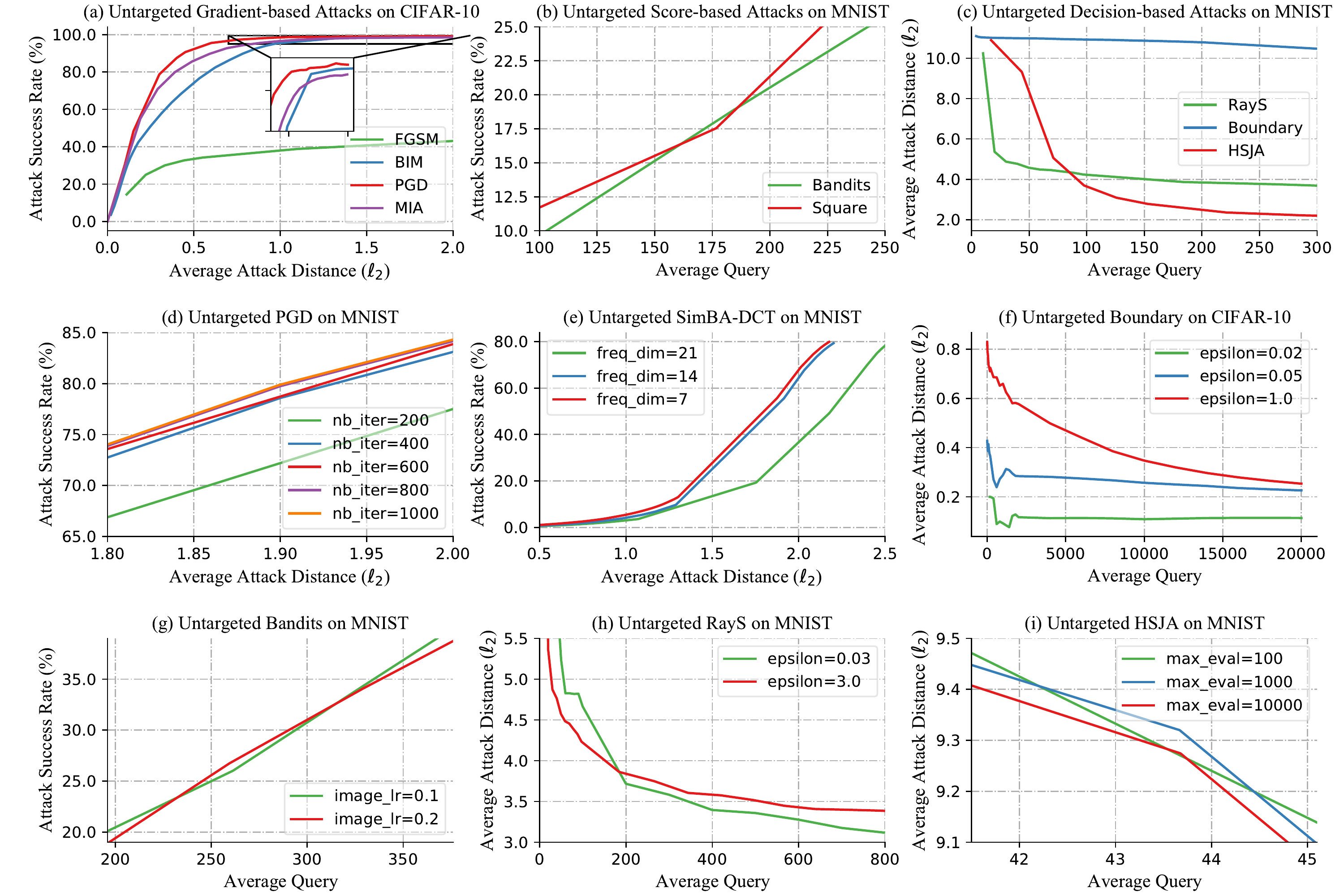}
\caption{The discrepancy for evaluating the performance of adversarial attacks. (a, b, c): the discrepancy in point-wise comparisons. (d, e, f): the discrepancy in parameter choosing. (g, h, i): the discrepancy in combining the results. For (a, b, c), we test lots of parameters for all adversaries and choose the best performance to plot the figure. In (b, g), we fix the perturbation $L_2$ distance as 3 $\pm$ 0.1.  In (c, f, h, i), we fix the attack success rate as 99.5 $\pm$ 0.5  \%. These discrepancies exist in the evaluation of gradient-based attacks, score-based attacks, and decision-based attacks. The parameters showed in this figure are corresponded to the libraries we used, see sup. for detail.}
\label{fig_intro}
\end{figure*}

However, there is a common problem in current comparison methods that we sometimes find the discrepancy in different studies, although they both get correct experimental results. For example, adversary A performs better than adversary B in a study. But in another study, we get the opposite conclusion. We find such discrepancies are caused by the following factors after analyzing the experiment results of 10 attack methods.

First, as shown in Figure~\ref{fig_intro}, (a), (b) and (c) shows the discrepancy among gradient-based, score-based and decision-based attacks, respectively. (a) plots $L_2$-ASR curve of gradient-based attacks on CIFAR-10. However, Figure~\ref{fig_intro} (a) shows $L_2$-ASR curves of Basic Iterative Method (BIM) and Momentum Iterative Attack (MIA) have intersection near the ASR of 100\%. So if we compare the performance of these methods by sampling the point before the intersection, MIA achieves better results in untargeted attacks on CIFAR-10. If we sample the point after the intersection, we will get the opposite conclusion. Therefore, it may lead to discrepancies in the evaluation. Besides, such discrepancy is also found in Figure~\ref{fig_intro} (b) and (c). Considering the diversity of adversaries, this problem will not only appear in classic methods but also in future attacks.

Second, in Figure~\ref{fig_intro} (d), (e) and (f), the parameter choosing do affect the performance evaluation for adversaries. Usually, we want to find the best parameter to show the performance of adversaries. We suggest the parameter choosing should be stated clearly in experiments to reduce the discrepancy. In this case, some adversaries can easily achieve their best performance. However, for some other adversaries, it is hard to say which set of parameters is the best. As shown in Figure~\ref{fig_intro} (g), (h) and (i), sometimes different parameters can achieve better performance under different situations. Should we evaluate the performance of attacks in a single set of well-chosen parameters for every attack? Or choose the upper bound of the results achieved by multiple sets of parameters for every attack to compare? Current studies lack this discussion, it may lead to discrepancies in the evaluation phase.

Third, in curve-based comparisons, like some current studies, it will lead to a new problem for us. There is no quantitative measurement or standardized curving method for curve-based comparisons. It also causes discrepancies. For example, it is hard to simply get a clear conclusion which method is the best in Figure~\ref{fig_intro} (c).

Due to the diversity of attacks, there is no `golden standard' to compare the performance of adversaries. It is hard to achieve a fair comparison without a fixed guideline. To address the above problems, we propose a Piece-wise Sampling Curving (PSC) framework to reduce the discrepancy in the comparison. Specifically, we use curves to depict the performance of attacks. Area Under the Curve (AUC) is then utilized to measure the overall performance in a specific perturbation distance range $\boldsymbol{\epsilon}$. Besides, for balancing the computational cost and evaluation effectiveness, the PSC framework has a resolution parameter $r$. The $\boldsymbol{\epsilon}$ then be divided into $r$ parts. For attackers, we should assume that they know how to maximize the performance of the adversaries, \eg \ how to select parameters in different perturbation distances. So, in this scenario, we choose one optimal results for multiple parameters in every part and plot the curve.

\begin{table*} [t]
	\caption{Survey of comparison methods used in current attack methods. One study might use multiple comparison methods. The experiments of PGD are focus on improving the robustness of the model by adversarial training, so we omit this method in this survey. Note that we just consider the experiments with comparison between different attack performance.}
	\label{current_standards}
	\centering
	\vspace{+0.5em}
		\begin{tabular}{|l|c|l|}
			\cline{1-3} 
			\multirow{10}{*}{Point-wise} 
			&&FGSM\cite{goodfellow6572explaining},JSMA\cite{papernot2016limitations},NES\cite{ilyas2018black},\\
			&  &ZOO~\cite{chen2017zoo}, UAN \cite{hayes2018learning}, SimBA\cite{guo2019simple},\\
			& & Targeted UAP\cite{hirano2020simple}, UAP\cite{moosavi2017universal},\\
			& single point &Rozsa\&Rudd\cite{rozsa2016adversarial},$\mathcal{N}$ATTACK \cite{li2019nattack},Pixel\cite{su2019one},\\
			& &   DeepFool\cite{moosavi2016deepfool},Boundary Attack\cite{brendel2017decision}, \\
			&&  SingularFool\cite{khrulkov2018art},Brendel\&Bethge \cite{NEURIPS2019_885fe656},\\	
			\cline{2-3} 
			&& BIM\cite{kurakin2016adversarial}, C\&W\cite{carlini2017towards}, DDN\cite{rony2019decoupling},\\
			&multiple points& Opt\cite{cheng2018query}, Sign-OPT~\cite{cheng2019sign}, RayS\cite{chen2020rays},\\
			&& Square Attack\cite{andriushchenko2020square}, HopSkipJump\cite{brendel2017decision}, \\
			&& AutoZOOM \cite{tu2019autozoom}\\
			\cline{1-3} 
			\multirow{3}{*}{Curve-based} 
			&without & BIM, UAP, Targeted UAP, \\
			&quantitative & SimBA, Opt, Sign-OPT, Square, AutoZOOM, $\mathcal{N}$ATTACK \\
			&measurement  & Brendel\&Bethge, HopSkipJump, RayS \\		
			\cline{1-3} 
		\end{tabular}
\end{table*}

\begin{figure*}[tbh]
\centering
\includegraphics[width=0.95\textwidth]{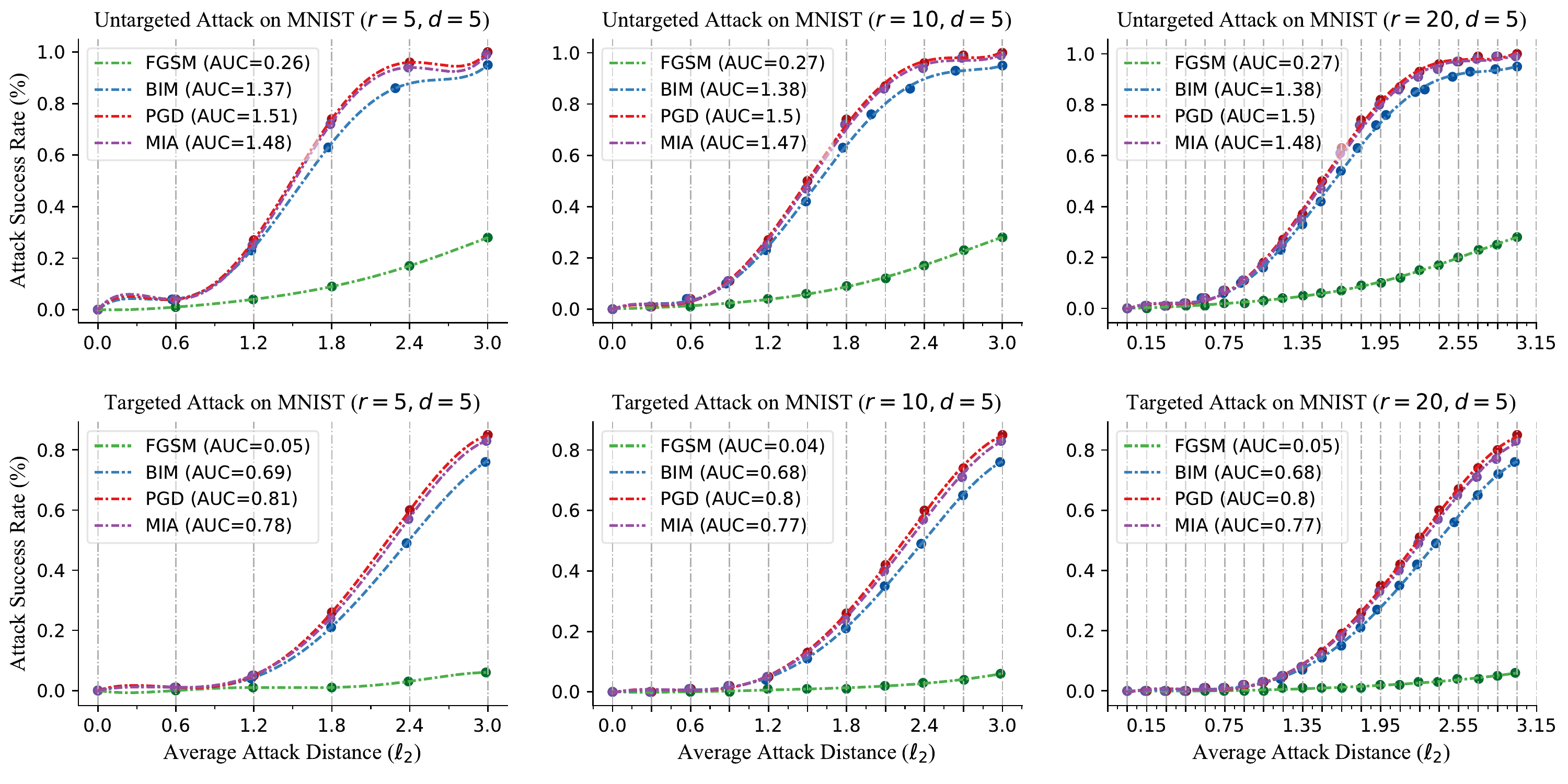}
\vspace{-0.5em}
\caption{Example of evaluating the performances on MNIST by PSC framework. The curves are fitted by 5-order functions. The resolution $r$ is 5, 10, and 20 from left to right, respectively. Top: untargeted attacks. Bottom: targeted attacks. The attacked model is a 4-layer Convolutional Neural Networks (CNNs). AUC: the higher is better. Because the performance of gradient-based attacks increases in the whole range, the best point in each part is near the boundary.}
\label{fig_introfitmn5}
\end{figure*}

Our contribution can be summarised as follows:
\begin{itemize}
    \item We analyze the discrepancy in current studies and build a unified framework to evaluate the performance of adversarial attacks. 

    \item According to the real-world attack scenario, we show some solutions to reduce the discrepancy, like using the upper bound of the evaluation performance generated by different adversary setting to plot the curve. 
    \item We propose the Piece-wise Sampling Curve (PSC) toolkit to reduce the discrepancy in the evaluation. We save our experiment results of current methods in the PSC. Users just need to upload their results to compare with current adversaries in a specific setting of PSC.  
\end{itemize}

\paragraph{Roadmap.} In section 2, we introduce the related adversarial attacks and comparison methods used in current studies. We show the principle of PSC in section 3. In section 4, we analyze the options of PSC and give some examples of how to handle the special methods. We conclude in section 5 and discuss disadvantage of the proposed PSC in section 6.
\section{Related work}

\subsection{Adversarial Attacks}
Current adversarial attacks can be divided into gradient-based attacks and query-based attacks. Specifically, each of these two categories can be further categorized into individual and universal adversarial attacks. The former need to craft every perturbation for every example, while the latter only need to construct a single perturbation for all benign examples. Current attack scenario includes the white-box setting where the adversary can access the internal information of the attacked models, and the black-box setting where the adversary only access the output returned by the attacked models.

\paragraph{Gradient-based attacks.}
Gradient-based attacks generate adversarial examples using the gradient information of the attacked models. For individual gradient-based attacks, FGSM~\cite{goodfellow6572explaining} computes the sign of gradient direction to modify the inputs, which showed the vulnerability of machine learning models. Then, BIM~\cite{kurakin2016adversarial} proposed an iterative version of FGSM and improved the performance. DeepFool~\cite{moosavi2016deepfool} is designed to optimize the direction for crossing the decision boundary, while C\&W~\cite{carlini2017towards} introduced a way to craft with minimal $L_p$ norm perturbations distance. To improve the robustness of models, PGD~\cite{madry2018towards} was proposed to generate adversarial examples for adversarial training. This method has become an important baseline for gradient-based attacks. Besides, DNN\cite{rony2019decoupling} is another efficient method to generate attacks, which decoupled the value and direction of the perturbation. To perform in the black-box setting, data-dependent attacks~\cite{papernot2017practical,tramer2016stealing} and data-free adversary~\cite{zhou2020dast} obtain a substitute model to generate adversarial examples for attacking the attacked models. For universal gradient-based attacks, data-dependant universal adversarial attacks~\cite{moosavi2017universal, khrulkov2018art,hayes2018learning,poursaeed2018generative} and data-free universal adversaries~\cite{mopuri2018ask,mopuri2017fast,mopuri2018generalizable} generate a image-agnostic perturbation.

\paragraph{Query-based attacks.}
For performing the adversarial attack directly in the black-box setting, query-based attacks are proposed to craft adversarial examples without the requirement of gradient information of the attacked models. They update their optimization step by sending numerous queries. This kind of attacks can be divided into score-based attacks that access the output probability returned by the attacked models, and decision-based (label-based) attacks having access to the inferred labels returned by the attacked models. For score-based attacks, adversaries~\cite{chen2017zoo,tu2019autozoom,IEM2018PriorCB} motivated by zeroth-order optimization estimate the gradient of the attacked model through the number of queries. Square Attack~\cite{andriushchenko2020square} and SimBA~\cite{guo2019simple} utilize a randomized search scheme. SimBA decides the direction of the perturbations based on the changes of the output probability. This method is simple yet very efficient. For decision-based attacks, Boundary Attack~\cite{brendel2017decision} first finds an adversarial example with large perturbation, then reduces the norm the perturbation. Also inspired by zeroth-order optimization, the advance version of Boundary Attack, HopSkipJump~\cite{brendel2017decision} , OPT~\cite{cheng2018query} and Sign-OPT~\cite{cheng2019sign} improves the query efficiency. For universal decision-based attacks, DUAttack~\cite{wu2020decision} construct the image-agnostic perturbation based on the randomized search scheme with momentum.

\subsection{Comparison Metrics}
Commonly, we need to measure the attack distance ($L_p$) and attack success rate (ASR). For query-based attacks, we need an extra measurement, namely the number of queries ($T$). $L_p(p=0,1,2,\infty)$ norm is usually used to represent the magnitude of perturbation. However, Sabour~\etal~\cite{sabour2015adversarial} concluded that $L_p$ norms are not the best available measurement to match with the human perception. Therefore, Rozsa~\etal\cite{rozsa2016adversarial} proposed Perceptual Adversarial Similarity Score (PASS) based on the Structural Similarity Index Measure (SSIM) for quantifying adversarial examples. However, the discrepancy cannot be reduced by changing one or two metric.

We investigate various studies and find current comparison methods for evaluating the performance of adversaries. As shown in Table~\ref{current_standards}, most studies use point-wise comparison, although we have proved that is not reliable. Besides, the curve-based comparison is just utilized to provide additional evidence for the effectiveness of attacks. They could alleviate the problem in point-wise comparison. But, because they lack quantitative measurement and guideline, curve-based comparison cannot appear independently.

How to use a unified framework to reduce the discrepancy in comparisons is open to date. We need to design a comparison standard to reduce the discrepancy, with quantitative measurement, and suitable for as many attacks as possible.
\begin{figure*}[tbh]
\centering
\includegraphics[width=0.95\textwidth]{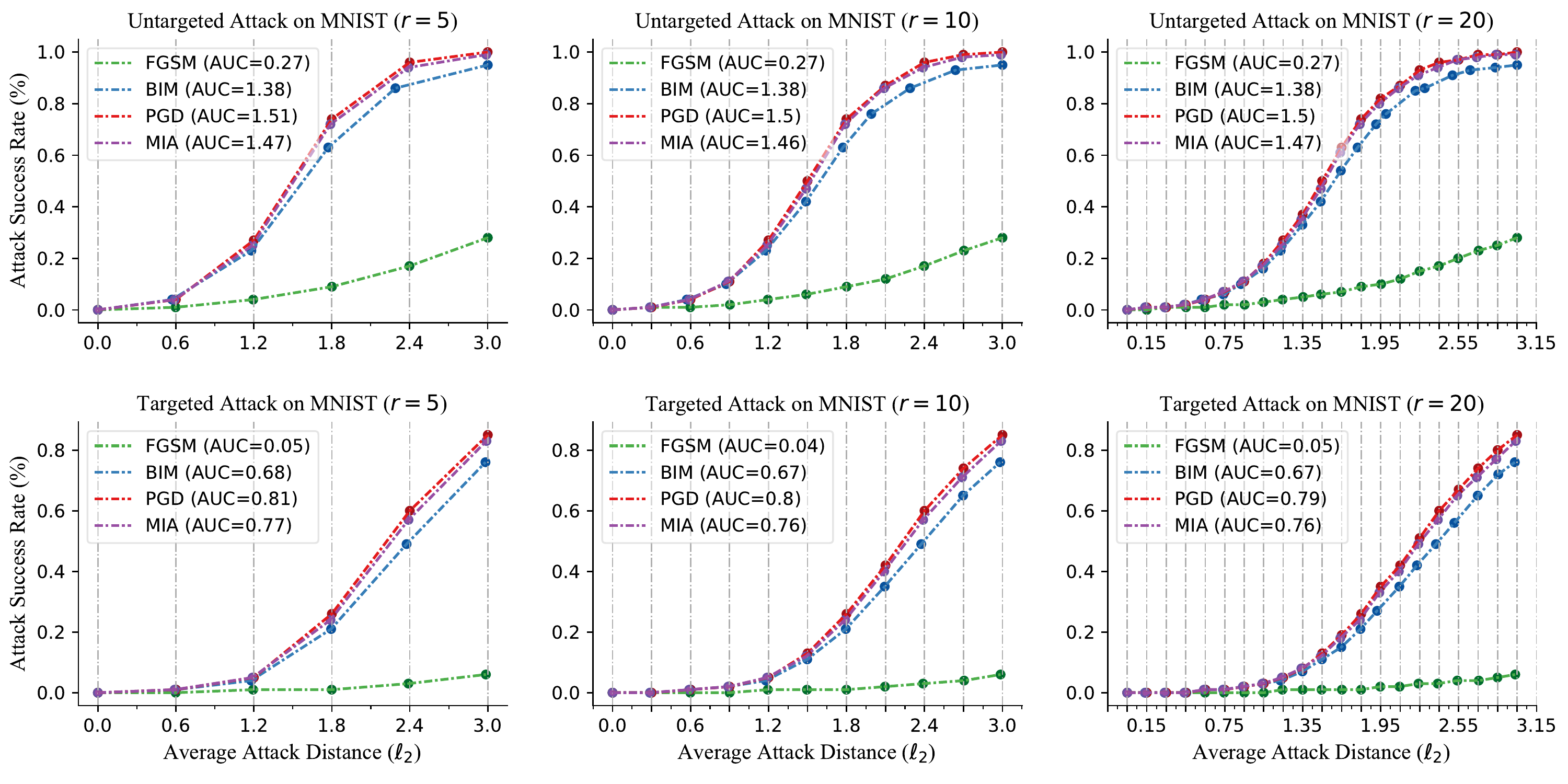}
\caption{Example of evaluating the performances of gradient-based attacks on MNIST by PSC framework. The curves are drew by straightly connecting the points. The attacked model is a 4-layer Convolutional Neural Networks (CNNs). AUC: the higher is better.}
\label{fig_introplymn}
\end{figure*}
\section{Piece-wise Sampling Curving Method}

\subsection{Scenario}
In this study, we assume attackers are capable to obtain the best performance of their methods in different $L_p$ norm distances. For example, an attacker wants to fool the autonomous intrusion detection system. When the system will be supervised by security staff, the attacker needs to limit the perturbation distance to generate imperceptible adversarial examples. When the system is running unattended, the limitation can be relaxed. In these two different situations, the attacker can test their methods in the simulation system. Then, they know the optimal parameters for different perturbation distance. For attack methods, the adversary has one optimal setting for the whole target perturbation distance range, which can be considered as a `simple method'. The opposite is called a 'complex method'.

In this case, to evaluate the potential maximum performance of the attack method, we need to test multiple sets of parameters to estimate the performance upper bound for the `complex method'.

\subsection{PSC Framework}
The proposed PSC framework has three steps, includes choosing a perturbation $L_p$ norm range. dividing the range, and curving. Besides, PSC has 4 parameters, includes comparison range $\boldsymbol{\epsilon}$, resolution $r$, order of fitting function $d$. 

In this subsection, for showing the procedure of PSC, we evaluate 4 gradient-based attack methods in the white-box setting, which is shown in Figure~\ref{fig_introfitmn5}.

\paragraph{Choosing a comparison range.} The PSC method compares different adversaries in a specific comparison range $\boldsymbol{\epsilon} = [\epsilon_{l}, \epsilon_{u}]$. We take the white-box attacks for an example, in Figure \ref{fig_introfitmn5}, the $\boldsymbol{\epsilon} = [0, 3.0]$. We have proof that one adversary may not perform best in the whole perturbation distance range. Therefore, for presenting the results, we should state the specific distance range clearly in comparison discussion. In some scenarios, attackers need to generate adversarial examples with small perturbations, but the others are opposites. So we need to evaluate the adversaries based on the target scenario. Also, we can test the adversary in a large range, and state clearly that the adversary performs well in which sub-range.
\begin{figure*}[t]
\centering
\includegraphics[width=0.8\textwidth]{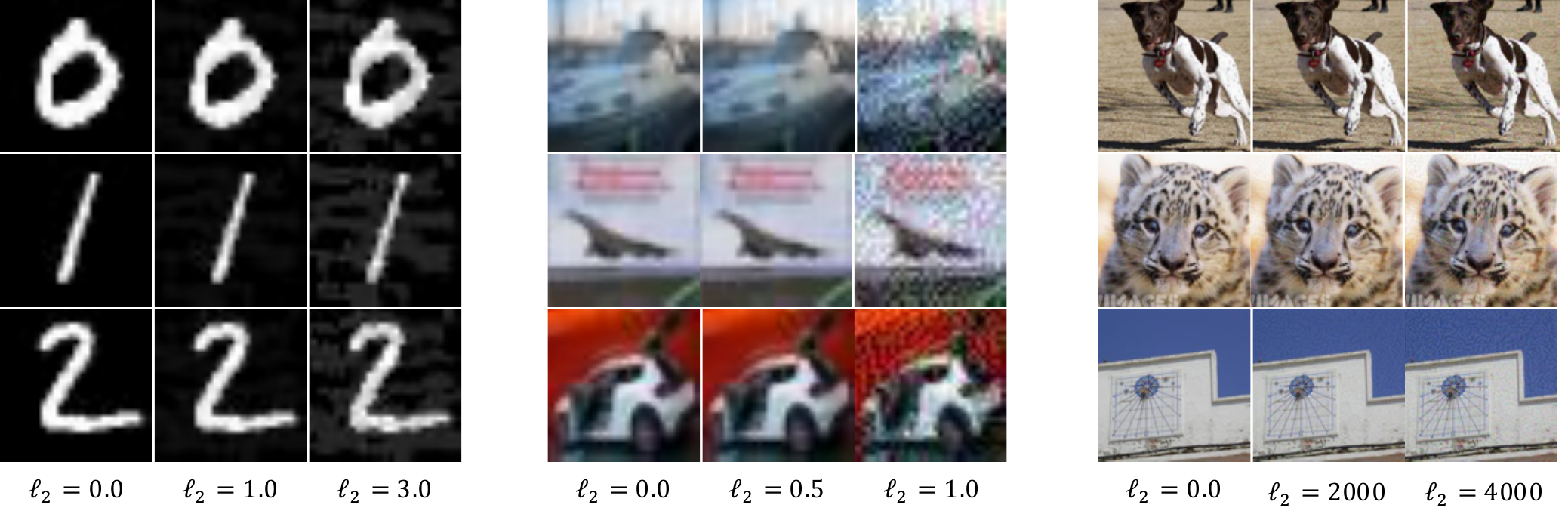}
\vspace{-0.5em}
\caption{Visualization of different perturbation $L_2$ distance. From left to rightL: MNIST, CIFAR-10, and ImageNet. }
\label{fig_vision}
\end{figure*}
\paragraph{Dividing the range.} For plotting the performance in a range $\boldsymbol{\epsilon}$, the most precise way is to sample points as much as possible. However, some adversaries cannot return a result in a short time on some large scale datasets. Therefore, for comparing the attack methods equally, we set a resolution parameter $r$ in PSC. It means the range $\boldsymbol{\epsilon}$ should be divided into $r$ part. In Figure \ref{fig_introfitmn5}, the resolutions are 5, 10, 20 from the left to right, respectively. Although the curve with a higher resolution might be more precise to depict the performance of attacks, it will cost more computational resources. We sample one best point for each part. We define the best point $x_n$ in one part as: 
\begin{align}
\label{formular:sampling}
\begin{split}
x_n =
\begin{cases} 
\text{max} (\bf{p}),  & \text{y-axis is ASR} \\
\text{min} (\bf{p}),  & \text{y-axis is distance or query} 
\end{cases} \\
\text{Subject to: } \mathbf{p} = \{ p_1, p_2, \cdots, p_i \}, 
\end{split}
\end{align}
where the $x_n$ denotes the best points in $n$-th part. Attacker method can be tested by multiple sets of parameters to obtain the $\bf{p}$, which is the set lied on $n$-th distance sub-range. When the y-axis of the curve is the attack success rate, the highest point achieves the best performances. When the y-axis is the perturbation distance and the number of queries, the lowest point achieves the best performances.

\begin{figure*}[t]
\centering
\includegraphics[width=0.88\textwidth]{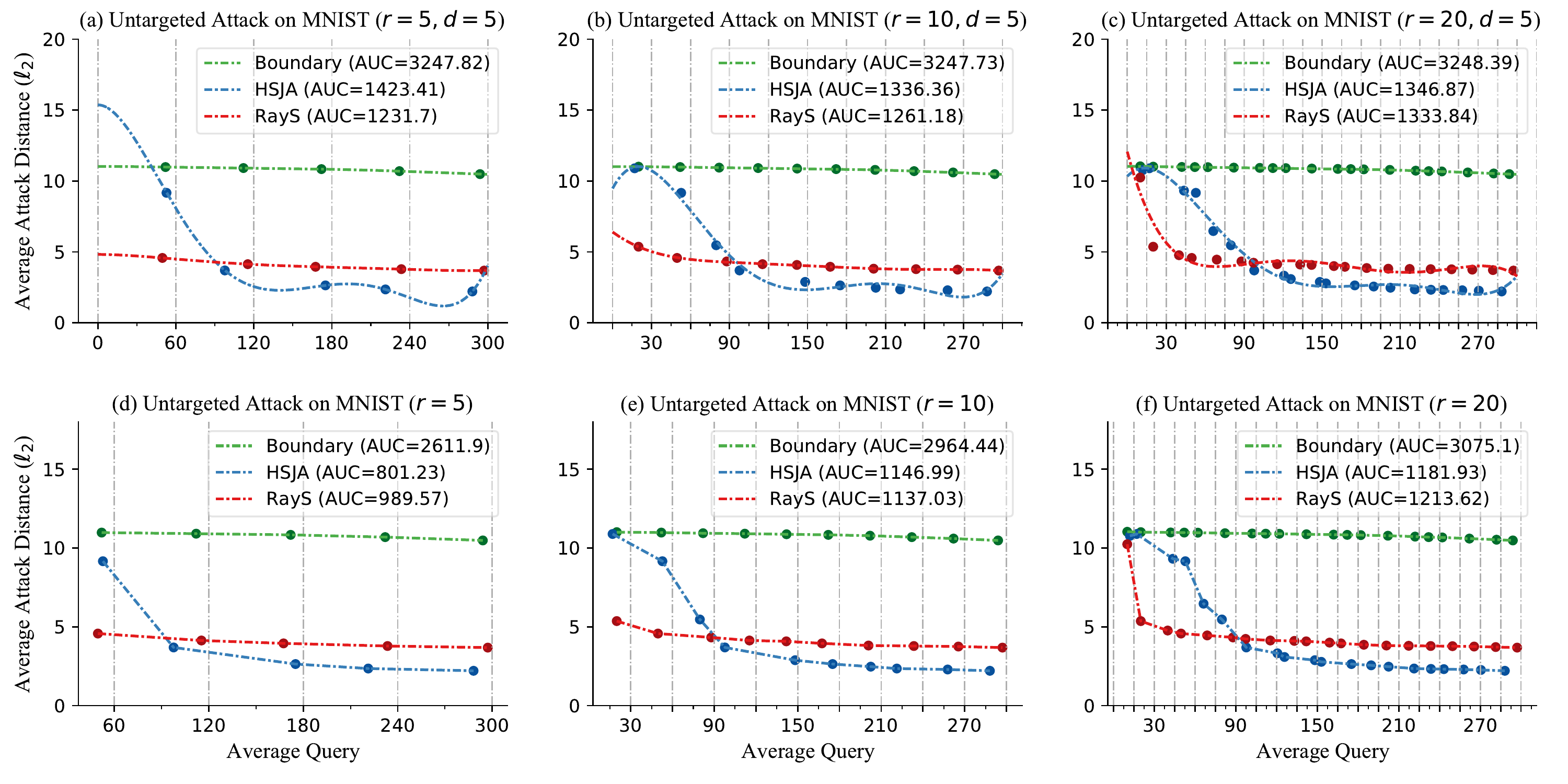}
\vspace{-0.5em}
\caption{Example of evaluating the performances of decision-based attacks on MNIST by PSC framework. The curves are generated by fitting 5-order functions and straightly connecting the points in the top and bottom of this figure, respectively. The resolution $r$ is 5, 10, and 20 from left to right, respectively. AUC: the lower is better.}
\label{fig_dbfitmn}
\end{figure*}

\paragraph{Curving.} After obtaining the sampling points, the PSC will use the fitting function to plot the figure. The order of a fitting function is $d, d > 0 $. In our experiments, the best order is 5. For fitting the sampling points, the function can be formulated as follows:
\begin{align}
\label{formular:fitting}
\begin{split}
f(\mathbf{x}) = \sum_{i=0}^d \alpha_i\mathbf{x}^i, d > 0\\
\end{split}
\end{align}
where $(\mathbf{x}, f(\mathbf{x}))$ is the set of sampling points. Therefore, to solve the $\boldsymbol{\alpha} = \{\alpha_1, \cdots, \alpha_d \}$, the resolution $r$ must larger than $d$. When the $d$ is set to be 0, we just connect the sampling points with straight lines, which is shown in Figure \ref{fig_introplymn}. We suggest that when the resolution $r$ is large, the order $d$ can be set to a large number.

\begin{figure*}[tbh]
\centering
\includegraphics[width=0.94\textwidth]{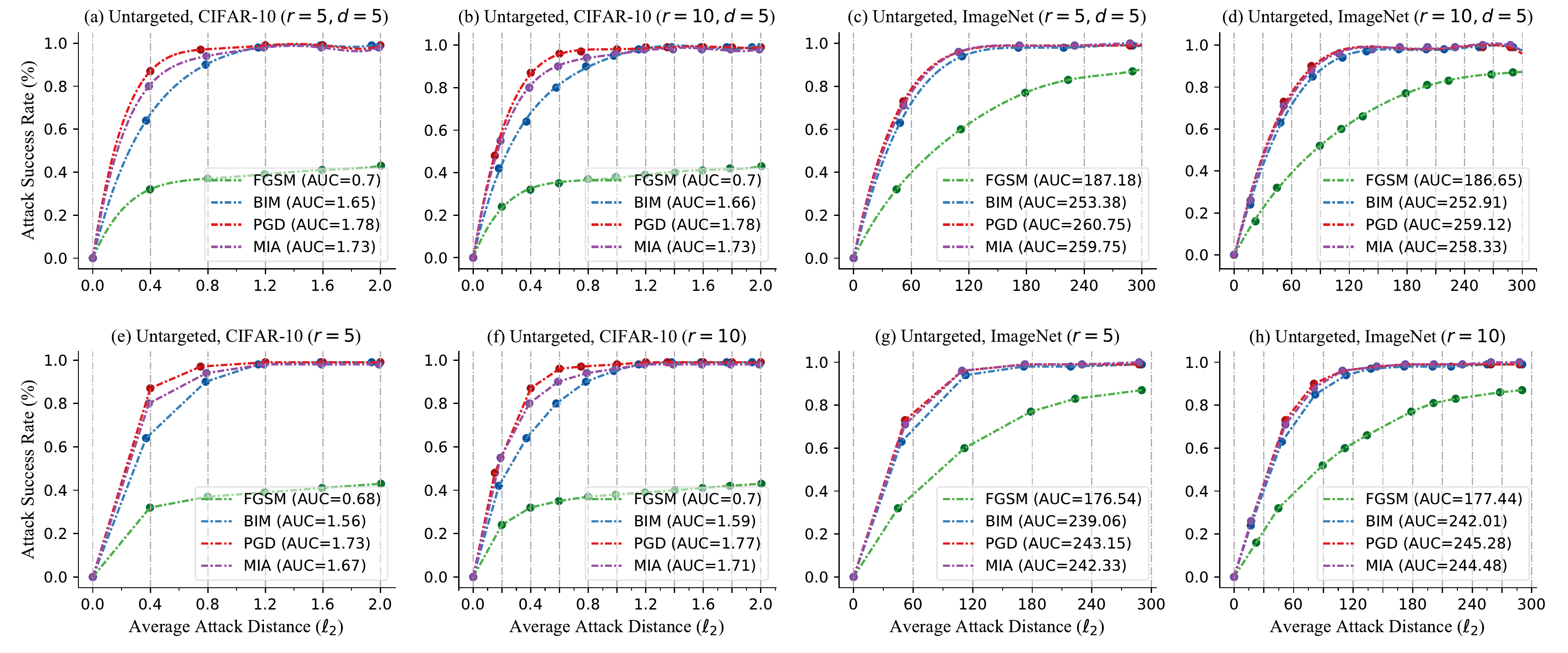}
\vspace{-0.5em}
\caption{Example of evaluating the performances of gradient-based attacks on CIFAR-10 and ImageNet by PSC framework. The curves are generated by fitting 5-order functions and straightly connecting the points in the top and bottom of this figure, respectively. AUC: the higher is better.}
\label{fig_gbciim}
\end{figure*}

\paragraph{Calculating AUC.} When one adversary exceeds others in the whole perturbation distance range $\boldsymbol{\epsilon}$, the problem will be simple. However, when curves cross in the range $\boldsymbol{\epsilon}$, we use the Area Under the Curve (AUC) for comparing the adversaries comprehensively. Compared with the evaluation using single or several points, our methods can reduce the discrepancy significantly. The procedure of our PSC is shown as Algorithm \ref{alg:PSC}.

\begin{algorithm}[tb]
   \caption{Piece-wise Sampling Curving framework}
   \label{alg:PSC}
\begin{algorithmic}

   \STATE {\bfseries Input:} Sampling points $\mathbf{x} = \{ x_1, \cdots, x_r \} $, resolution $r$, order $d$, comparison range $\boldsymbol{\epsilon} = [\epsilon_{l}, \epsilon_{u}]$
   \STATE {\bfseries Procedure:}
   \STATE \qquad Divide the $\boldsymbol{\epsilon}$ into $r$ parts.
   \STATE \qquad Sampling the best point in each part.
   \STATE \qquad Fitting the Equation (\ref{formular:fitting}) by Sampling points $\mathbf{x}$.
   \STATE \qquad Calculating the AUC.
   \STATE {\bfseries Output:} curve and AUC
\end{algorithmic}
\end{algorithm}

\section{Experimental Analysis for PSC}
In section 3, we introduce the PSC and show some examples for comparing white-box attacks. However, adversaries are very diverse. It is hard to simply use the PSC in some special cases. In this section, we will show how to evaluate the performance of query-based attacks based on our PSC framework. Besides, we will explain some basic features for our PSC framework. Before that, we visualize some specific perturbation $L_p$ norm distances on 
MNIST, CIFAR-10 \cite{krizhevsky2009learning}, ImageNet\cite{deng2009imagenet} datasets in Figure \ref{fig_vision}. 

\paragraph{Setting.} We utilize a 4-layer CNNs, VGG-16, and VGG-16 as the attacked model on MNIST, CIFAR-10 and ImageNet datasets, respectively. Note that the scale of pixels on MNIST, CIFAR-10, and ImageNet are [0, 1], [0, 1], [0, 255], respectively. We use Advertorch \cite{ding2019advertorch}, Adversarial Robustness Toolkit \cite{nicolae2018adversarial}, and Foolbox \cite{rauber2020foolbox} to re-implement adversaries.

\subsection{Experimental Analysis for Query-based Attacks.}
Compared with gradient-based attacks, the measurements of query-based attacks have more dimensions, includes the attack success rate, perturbation distance, and number of queries. Therefore, for comparing the query-based attacks, we also need to consider the evaluation with more metrics. 

\paragraph{Dimension.} For query-based attacks like Bandits Attack, and Square Attack, their measurements do have three dimensions. But for some decision-based attacks like Boundary Attack and HopSkipJumpAttack, they actually have two measurements. Because they initially create the adversarial examples with near 100\% attack success rate and large perturbation distances. Then, they try to find perturbations with the smallest distance while maintaining the attack success rate. Although they do have some extreme cases that the attack success rate is much less than 100\%. All the results used for comparison should be obtained when the attack method works well. Therefore, for adversaries like Boundary and HopSkipJumpAttack, we just need to consider the perturbation distance and number of queries as the attack success rate is fixed.

\begin{figure}[t]
\centering
\includegraphics[width=0.44\textwidth]{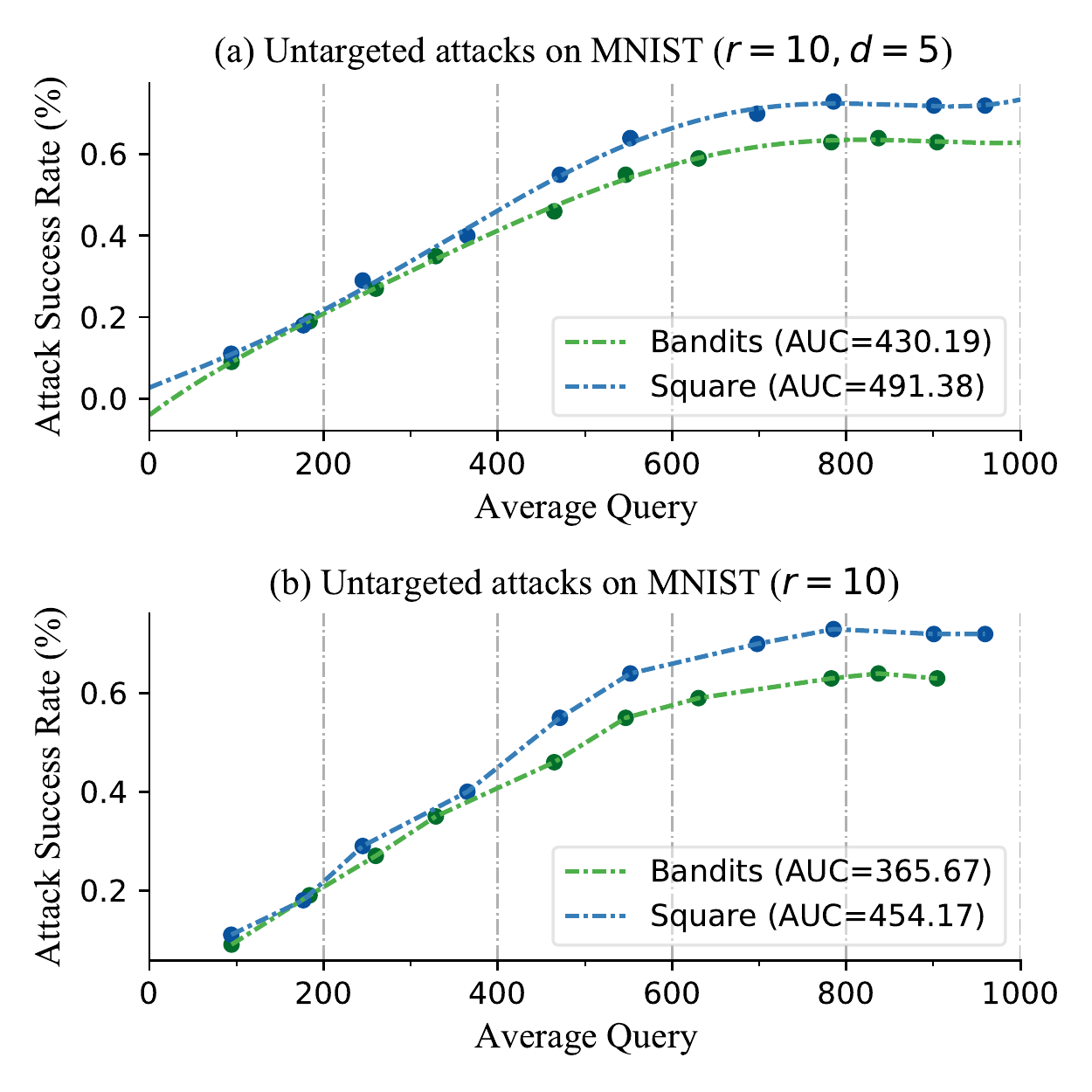}
\vspace{-1em}
\caption{Example of evaluating the performances of score-based attacks on MNIST by PSC framework. We fix the $L_2$ distance as 3.0 $\pm$ 0.1. The curves are generated by fitting 5-order functions and straightly connecting the points in the top and bottom of this figure, respectively. AUC: the higher is better.}
\label{fig_sb}
\end{figure}

\begin{figure*}[t]
\centering
\includegraphics[width=0.89\textwidth]{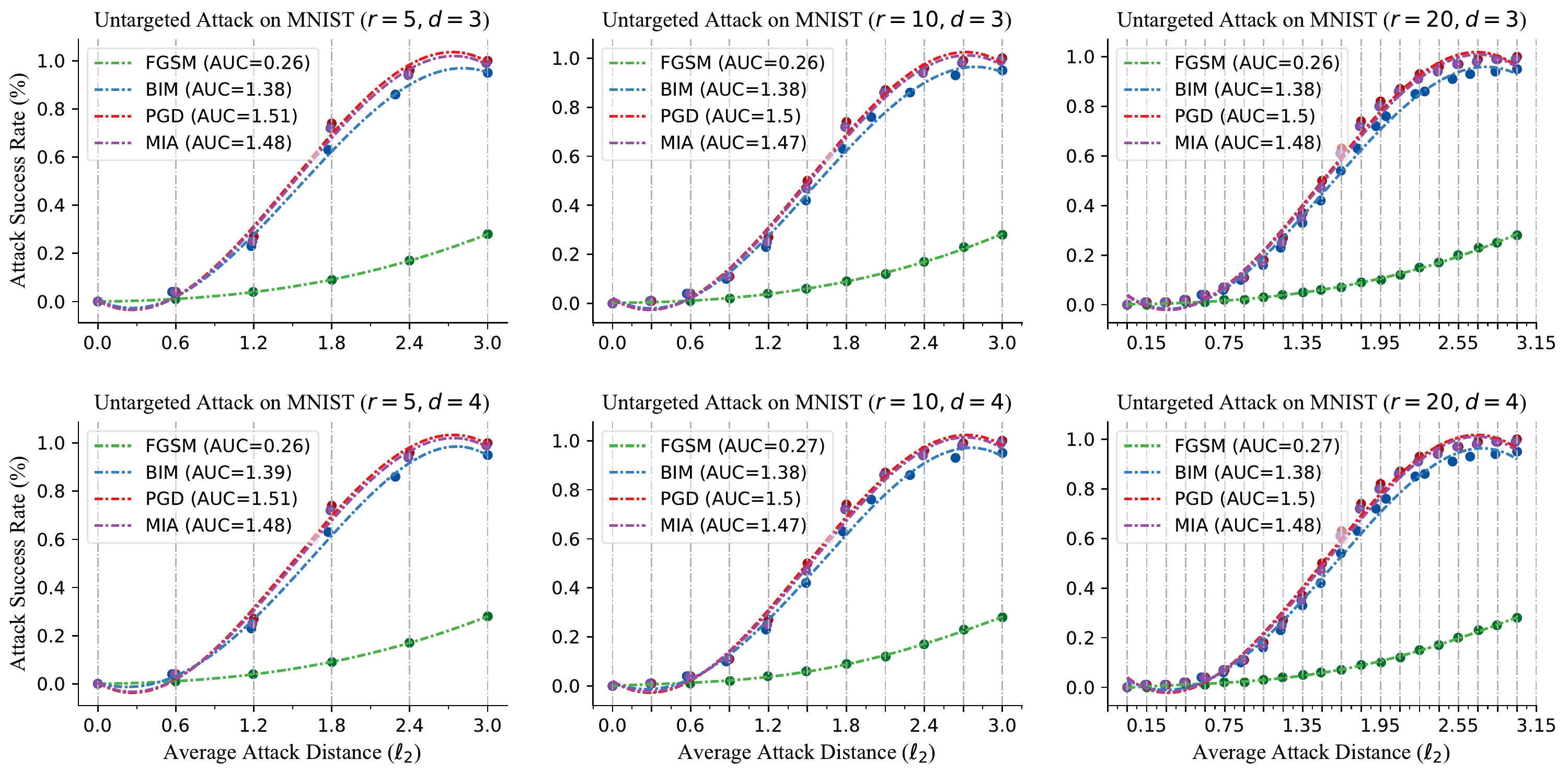}
\caption{Example of evaluating the performances of gradient-based attacks on MNIST by PSC framework. The top and bottom curves are fitted by 3-order and 4-order functions, respectively. AUC: the higher is better.}
\label{fig_order}
\end{figure*}

\paragraph{Example for 2-dimension comparisons.} For decision-based attacks, Figure~\ref{fig_dbfitmn} shows an example to compare the performance of Boundary Attack, HopSkipJumpAttack, and Rays Attack. This comparison is similar to the Figure~\ref{fig_introfitmn5}. Compared with Figure~\ref{fig_introfitmn5}, the method with less AUC performs better than others.

\paragraph{Example for 3-dimension comparisons.} For score-based attacks, Figure~\ref{fig_sb} shows an example to compare the performance of Bandits Attack and Square Attack. We fix the perturbation $L_2$ distance. Besides, for example, in some special scenarios, if the number of queries is not be limited in the attacked system, we just need to consider the attack success rate and the perturbation distance. For the systems do not limit the perturbation distance, we can just consider the attack success rate and queries. Therefore, in these cases, they are actually the 2-dimension comparison. For space limitation, we omit to show some examples in this scenario, the usage of PSC is very similar to the previous cases.

\paragraph{Example for CIFAR-10 and ImageNet datasets.} We also show examples of how to evaluate the performance of gradient-based attacks on CIFAR-10 and ImageNet, which is shown in Figure~\ref{fig_gbciim}.
 
\subsection{Suggestions for Parameters}

\paragraph{The trade-off between the cost and effectiveness.} According to the results of Figure ~\ref{fig_introfitmn5} and ~\ref{fig_introplymn}, we find the curve will be more smooth if we set a large resolution $r$. However, sampling more points means more computational costs. It is necessary that balancing computational costs and evaluation effectiveness. For suggestion, we can set a small resolution first. When we find we cannot obtain a good curve in current setting, we then increase the resolution until getting a well-fitting curve.

\paragraph{Order of Fitting.} We test different order of fitting for PSC. Figure~\ref{fig_introfitmn5} and \ref{fig_order} show when the order is set to be less than 5, it will cause under-fitting, which the curves cross the absolute upper (100\%) and lower (0\%) bounds in the fitting. When the order is set to be 5, this problem disappears. Therefore, we choose the 5 as the default order. However, for solving the fitting function, the $d$ must less than the $r$. Besides, we also offer options that plotting curves by straightly connecting the points. We recommend that using $d = 0$ when the resolution $r \le 5$.

\section{Conclusion}
In this study, We propose a PSC framework for reducing the discrepancy in the evaluation of adversarial attacks. There is no unified evaluation framework in current studies. The PSC has four main steps, all steps can be standardized. We also provide options for PSC framework, they can balance the computational cost and evaluation effectiveness. For some special attack methods, We give examples of how to evaluate their performances. 

Besides, we organized our experiment data and codes to build a PSC toolkit. It offers experiment results of previous methods, users just need to upload their results under a specific setting. we believe it will standardize and speed up the equal comparison of adversarial attacks. 

\section{Future Perspectives}
In the future, we plan to build more effective methods to evaluate the performance of attacks with less computational costs. In summarize, the PSC has two main problems. 

First, in the current PSC framework, the evaluate effectiveness relies on the number of sampling points. We cannot increase the number of samples while reducing the total consumption time. Second, it is hard to obtain which order of fitting function is the best. To address this problem, we must try different orders in practice.
\bibliography{example_paper}
\bibliographystyle{icml2021}

\end{document}